\documentclass[letterpaper]{article}
\usepackage{aaai}
\usepackage{times}
\usepackage{helvet}
\usepackage{courier}
\usepackage{graphicx}
\usepackage{epsfig}
\usepackage{algorithm}
\usepackage{algorithmic}
\usepackage{amssymb}
\usepackage{amsthm,amssymb}
\usepackage{amsmath}
\usepackage{epstopdf}

\newcommand{\tool}[1]{$\textsf{#1}$}

\begin{document}
% Pages 8 + 1 (references) - Abstract (9 nov) - Paper (16 nov)
\title{Handling PDDL3.0 State Trajectory Constraints with Temporal Landmarks}
\author{Eliseo Marzal, Mohannad Babli, Eva Onaindia, Laura Sebastia \\
Universitat Polit{\`{e}}cnica de Val{\`{e}}ncia, Camino de Vera s/n\\ E46022-Valencia (Spain)\\
\{emarzal, mobab, onaindia, lstarin\}@dsic.upv.es\\
}
\maketitle

\begin{abstract}
\begin{quote}
Temporal landmarks have been proved to be a helpful mechanism to deal with temporal planning problems, specifically to improve planners performance and handle problems with deadline constraints. In this paper, we show the strength of using temporal landmarks to handle the state trajectory constraints of PDDL3.0. We analyze the formalism of \tool{TempLM}, a temporal planner particularly aimed at solving planning problems with deadlines, and we present a detailed study that exploits the underlying temporal landmark-based mechanism of \tool{TempLM} for representing and reasoning with trajectory constraints.
\end{quote}
\end{abstract}

%Abstract + Introduction = 1 p.
%Foundations + Example = 1 p.
%Overview + Landmarks STRIPS = 0.5 p.
%Temporal Landmarks + Intervals = 2.5 p.
%Search = 2.5 p.
%Experiments + Conclusions = 1 p.
%References = 1 p.

\section{Introduction}

In planning, a \emph{landmark} is a fact that must be true in any solution plan. Since the influential work presented in \cite{hoffmann04} on the use of landmarks in planning, there have been multiple investigations that exploit landmarks for cost-optimal planning \cite{helmert09,karpas09}, satisficing planning \cite{richter10} or more recently on goal recognition \cite{Frage17}.

Although the use of landmarks in temporal contexts has been less explored, there are two works that address the exploitation of temporal landmarks in planning. In \cite{Karpas15}, the authors define temporal fact landmarks (facts that must hold between two time points) and temporal action landmarks, which state that some event (the start or end of an action) must occur at some time point. The temporal information is captured in a Simple Temporal Network \cite{Dechter:1991} over the symbolic time points associated with each landmark. This approach is aimed at deriving temporal landmarks and constraints from planning problems and integrating them into domain-independent temporal planners in order to improve their performance. Overall, the results reveal that there is some benefit from using temporal landmarks in concurrent temporal planning problems like the \tool{TMS} domain and in some non-temporally expressive domains in terms of solution quality.

Differently, the approach presented in \cite{MarzalSO14,MarzalSO16}, called \tool{TempLM}, studies the use of temporal landmarks for solving planning problems with deadline constraints. In this approach, a temporal landmark is defined as a fact that must be achieved in a solution plan to satisfy the problem deadline constraints. \tool{TempLM} discovers the causal (non-temporal) landmarks of the problem and then exploits deadlines to infer new (temporal) landmarks that must be accomplished to meet the deadlines. The limitation of \tool{TempLM} is that it relies on the specification of deadlines for the goal propositions of the problem and it requires an upper time bound for the plan, which is automatically derived from the problem deadlines. This way, when there are no deadlines in the problem, a sufficiently large estimated value must be provided as upper bound of the plan. \tool{TempLM} shows an excellent performance in overconstrained problems with tight deadlines, which would clearly degrade with larger plan upper bounds.

Nonetheless, the key contribution of \tool{TempLM} is its internal representation of temporal landmarks, which can be easily used to handle all the state trajectory constraints defined in PDDL3.0 \cite{GereviniHLSD09}. In this paper, we show that the interval representation of temporal landmarks of \tool{TempLM}, along with the constraints defined between landmarks and their intervals, constitute a very suitable framework for representing and reasoning with trajectory constraints. The next section summarizes the main features of \tool{TempLM}, highlighting the representation of a landmark and the propagation of temporal constraints. The following section presents the 10 modal operators that specify the state trajectory constraints in PDDL3.0; for each operator, we show the corresponding landmark representation and the set of constraints that would be needed in \tool{TempLM} in order to account for the constraint. Subsequently, we present an illustrative example that shows the powerful inference engine of \tool{TempLM} when handling trajectory constraints. In the last section, we discuss the advantages and limitations of \tool{TempLM} and we stress the utilization of temporal landmarks for representing other constraints in temporal planning.

\section{Overview of \tool{TempLM}}
\label{sec:TempLM_background}

\tool{TempLM} is a framework specifically aimed at solving temporal planning problems with deadline constraints \cite{MarzalSO14,MarzalSO16}. It assumes a subset of the semantics of the temporal model of PDDL2.1 \cite{fox03}, the Time-Initial Literals (TILs) defined in PDDL2.2 \cite{hoffmann:edelkamp:jair-05} as well as the \texttt{within} constraint introduced in PDDL3.0 \cite{GereviniHLSD09}.

A {\bf temporal planning problem with deadline constraints} is a tuple $\mathcal{P}=\big<P,O,I,G,D\big>$, where $P$ is a set of propositions, $I$ is the initial state, $O$ is a set of durative actions in PDDL2.1, $I$ is the initial state, $G$ is a goal description and $D$ is a set of deadline constraints of the form $(p,t)$, denoting that proposition $p$ must be achieved within $t$ time units. A durative action $a \in O$ in PDDL2.1 (\cite{fox03}) is defined as a tuple $\langle dur(a),Cond(a), Eff(a) \rangle$ where $dur(a) \in \mathcal{R}^{+}$ is the duration of the action; $Cond(a)=SCond(a) \cup ECond(a) \cup Inv(a)$ (conditions to hold at the start, at the end or overall the duration of $a$); $Eff(a)=SEff(a) \cup EEff(a)$ (effects produced at the start or end of the execution of $a$).

A {\bf temporal plan $\Pi$} is a set of pairs $(a,t)$, where $a \in O$ and $t$ is the start execution time of $a$. For a given proposition $p$, $start(p)$ and $end(p)$ denote the time points when $p$ is asserted and deleted, respectively, by any action $a$ in $\Pi$. The duration (makespan) of a temporal plan $\Pi$ is $dur(\Pi)={\max\limits_{\forall (a, t) \in \Pi}}\big(t+dur(a)\big)$. Additionally, the upper bound of the temporal plan $\Pi$ is set as $T_{\Pi}= \max\limits_{t}(p,t) \; , \forall (p,t) \in D$.

\vspace{0.1cm}

\tool{TempLM} extracts first the non-temporal landmarks of a problem $\mathcal{P}$ \cite{hoffmann04} and then discovers a new set of (temporal) landmarks through the deadline constraints in $D$. A \textbf{temporal landmark} of a problem $\mathcal{P}$ is a proposition of $P$ that must hold in every plan that solves $\mathcal{P}$ in order to satisfy $D$. In this paper, we focus exclusively on describing the internal representation of temporal landmarks in \tool{TempLM}. We refer the reader to the works in \cite{MarzalSO14,MarzalSO16} for details of the landmark extraction process.

\subsection{Temporal Landmarks}
\label{sec:TL}
\begin{figure}[h]
%\begin{center}
{\centering
\includegraphics [width=8cm]{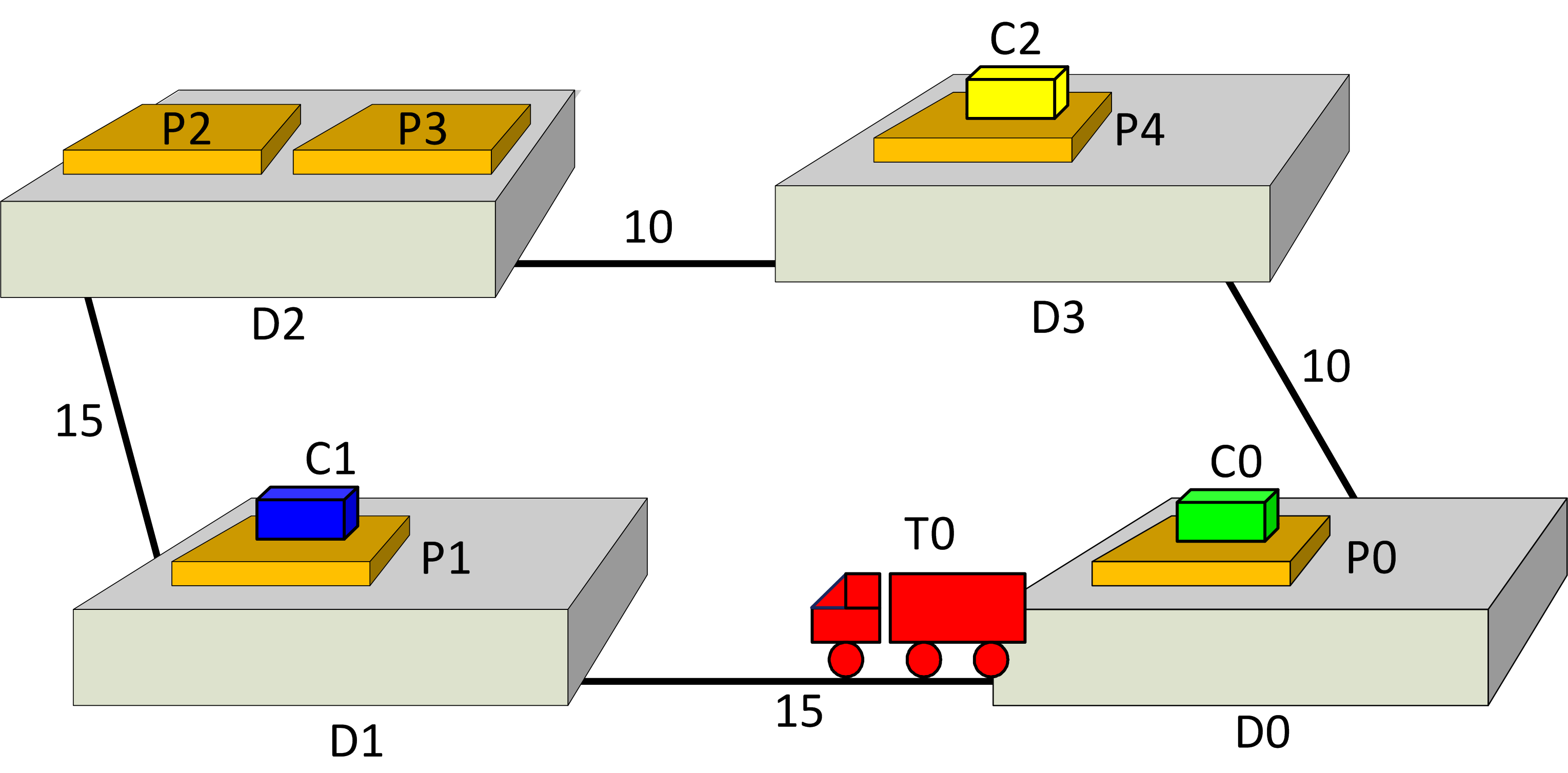}%[width=12cm]{./Figuras/esquema.png}
\caption{Illustrative example} \label{fig:depots}
}
%\end{center}
\end{figure}

We introduce a running example on the classical \texttt{depots} domain (Figure \ref{fig:depots}) in order to show the relevant information of the temporal landmarks. The scenario consists of a depot \texttt{D0}, where the pallet \texttt{P0} and the truck \texttt{T0} are located; a crate \texttt{C0} is in pallet \texttt{P0}.  There is also a distributor \texttt{D1} which contains create \texttt{C1} in pallet \texttt{P1}, and a distributor \texttt{D2} that contains the pallets \texttt{P2} and \texttt{P3}. Finally, the crate \texttt{C2} is in pallet \texttt{P4} located in distributor \texttt{D3}. The time a truck takes to travel between depots or distributors is shown on the edges. For simplicity, the hoists of the original domain have been eliminated: crates can be loaded into the truck as long as they are clear (have nothing on top) and can be unloaded from the truck to be put on top of another crate or onto a pallet.

Landmarks are characterized by their temporal intervals and relationships between them. Landmarks form a {\bf Temporal Landmarks Graph (TLG)}, a directed graph $G=(V,E)$ where $V$ are landmarks and $E$ is a set of ordering constraints of the form $l_i \prec_{n} l_j$ or $l_i \prec_{d} l_j$ that denote a \emph{necessary} or \emph{dependency} ordering, respectively, meaning that $l_i$  must happen before landmark $l_j$ in every solution plan.

\vspace{0.1cm}

Let's assume the goal of the problem is \texttt{((at C0 D2),40)}, which will be specified as  \texttt{(within 40 (at C0 D2))}. Then, \texttt{(at C0 D2)} is a temporal landmark and likewise all the propositions of the initial state and goal state. In this situation, a dependency ordering $\texttt{(at T0 D0)} \prec_{d} \texttt{(at T0 D2)}$ is established, which denotes that even though there are two possible ways of reaching the goal - through distributor \texttt{D1} or \texttt{D3}- \texttt{(at T0 D0)} must always be satisfied before \texttt{(at T0 D2)} in any solution plan. However, if the goal were \texttt{((at C0 D2),25)} then the TLG would contain the necessary ordering $\texttt{(at T0 D0)} \prec_{n} \texttt{(at T0 D3)}$ because in this case it is mandatory that truck \texttt{T0} travels through \texttt{D3} to reach the goal in time. Another interesting aspect is that since the surface which \texttt{C0} must be stacked on is not known, the propositions \texttt{(on C0 P2)} and \texttt{(on C0 P3)} are not landmarks.

Landmarks are also annotated with various {\em temporal intervals} that represent the validity of the corresponding temporal proposition \cite{marzal08}. Three types of intervals are identified:

\begin{itemize}
\item The \textbf{generation interval} of a landmark $l$ is denoted by $[min_g(l), max_g(l)]$. $min_g(l)$ represents the earliest time point when landmark $l$ can start in the plan. This value is determined by the time of the first proposition layer when $l$ appears in a Temporal Relaxed Planning Graph (TRPG). $max_g(l)$ represents the latest time point when $l$ must start in order to satisfy the deadlines $D$ of a problem $\mathcal{P}$ and it is initialized as $max_g(l)=T_{\Pi}$.
\item The {\bf validity interval} of a landmark $l$ is denoted by $[min_v(l), max_v(l)]$ and it represents the longest time that $l$ can hold in the plan. Initially, this interval is set as $min_v(l)=min_g(l)$ and $max_v(l)=T_{\Pi}$.
\item The {\bf necessity interval} of a landmark $l$ is denoted by $[min_n(l), max_n(l)]$ and it represents the set of time points when $l$ is required as a condition for an action to achieve other landmarks. Initially, $min_n(l)=min_g(l)$ and $max_n(l)=T_{\Pi}$.
\end{itemize}

Let us assume the \texttt{load} and \texttt{unload} actions in the example of Figure \ref{fig:depots} have a duration of two time units each; and that the problem goal is $g= \texttt{((at C0 D2),25)}$, being this the only deadline constraint of the problem. Figure \ref{fig:TLG_ini_goal25before} shows the initial TLG for this goal. Thus, $T_{\Pi}=25$ and the generation interval of $g$ is:

\begin{itemize}
\item $max_g(g)=25$ because the latest time at which $g$ must be generated in order to satisfy the deadline is 25
\item $min_g(g)=22$ because the first appearance of \texttt{(at C0 D2)} in the TRPG is at level 22: 20 (shortest route) + 2 (\texttt{unload}) (note that the first TRPG layer that contains the effects of \texttt{load} is at level 2).
\end{itemize}

For landmark $l_1= \texttt{(at T0 D0)}$ we have that $min_g(l_1)=0$ and $max_g(l_1)=25$; for landmark $l_2= \texttt{(at T0 D3)}$, $min_g(l_2)=10$ and $max_g(l_2)=25$; and for $l_3= \texttt{(at T0 D2)}$, $min_g(l_3)=20$ and $max_g(l_3)=25$. Likewise, the validity intervals would initially take on the same values as for the generation intervals.

\subsection{Propagation of temporal constraints}

Once the intervals of the temporal landmarks are initialized in the TLG, constraints are propagated and the landmark intervals are updated accordingly.

\vspace{0.1cm}

\textbf{\emph{Causal relationships}}. The ordering constraints $l_i \prec_{n} l_j$ or $l_i \prec_{d} l_j$ represent causal relationships, where $l_i \in Cond(a)$ and $l_j \in Eff(a')$ for two actions $a, a' \in O$. If $a=a'$ then it is a direct causal relationship represented by $l_i \prec_{n} l_j$. In any other case, $l_i \prec_{d} l_j$ represents an indirect causal relationship that involves more than one action. The necessary and dependency orderings are transitively propagated across the TLG creating further constraints. Particularly, for two landmarks involved in a causal relationship,  a separating temporal distance between the time point when $l_i$ is required and the time point when $l_j$ is needed is defined according to the duration of the action(s) involved in the causal transition. Restricting our attention to the simple case when $a=a'$, we have that: (a) $dist(l_i,l_j)=dur(a)$ if $l_i \in SCond(a)$ and $l_j \in EEff(a)$; (b) $dist(l_i,l_j)=\epsilon$ if $l_i \in ECond(a)$ and $l_j \in EAdd(a)$; and (c) $dist(l_i,l_j)=-dur(a)$ if $l_i \in ECond(a)$ and $l_j \in SAdd(a)$\footnote{The definition of $dist(l_i,l_j)$ is also dependent on the first or the last time when $l_i$ is required. For the sake of simplicity, we define a general concept of distance. More details on this can be found in \cite{MarzalSO14}.}.

\begin{figure}
%\begin{center}
{\centering
\epsfig{file=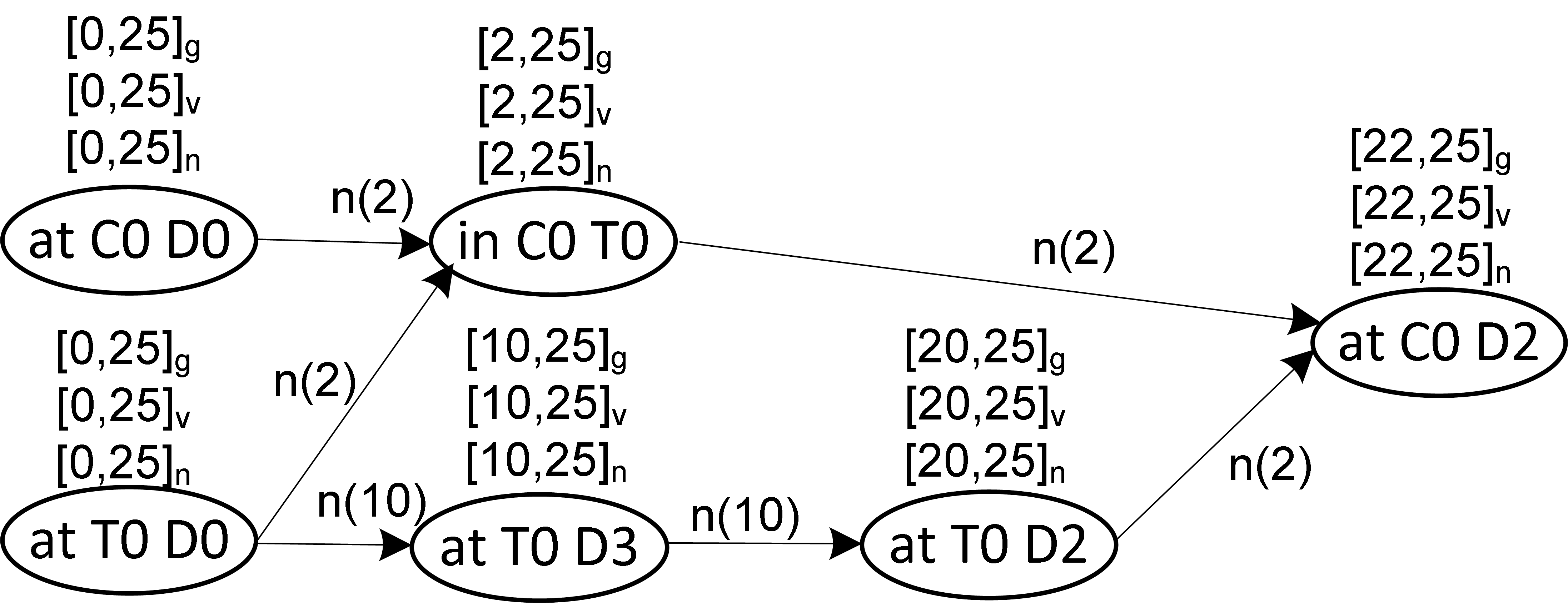, width=8cm}
\caption{Initial TLG for the goal {\small\texttt{(within 25 (at C0 D2))}}}
\label{fig:TLG_ini_goal25before}
}
%\end{center}
\end{figure}

\begin{figure}
%\begin{center}
{\centering
\epsfig{file=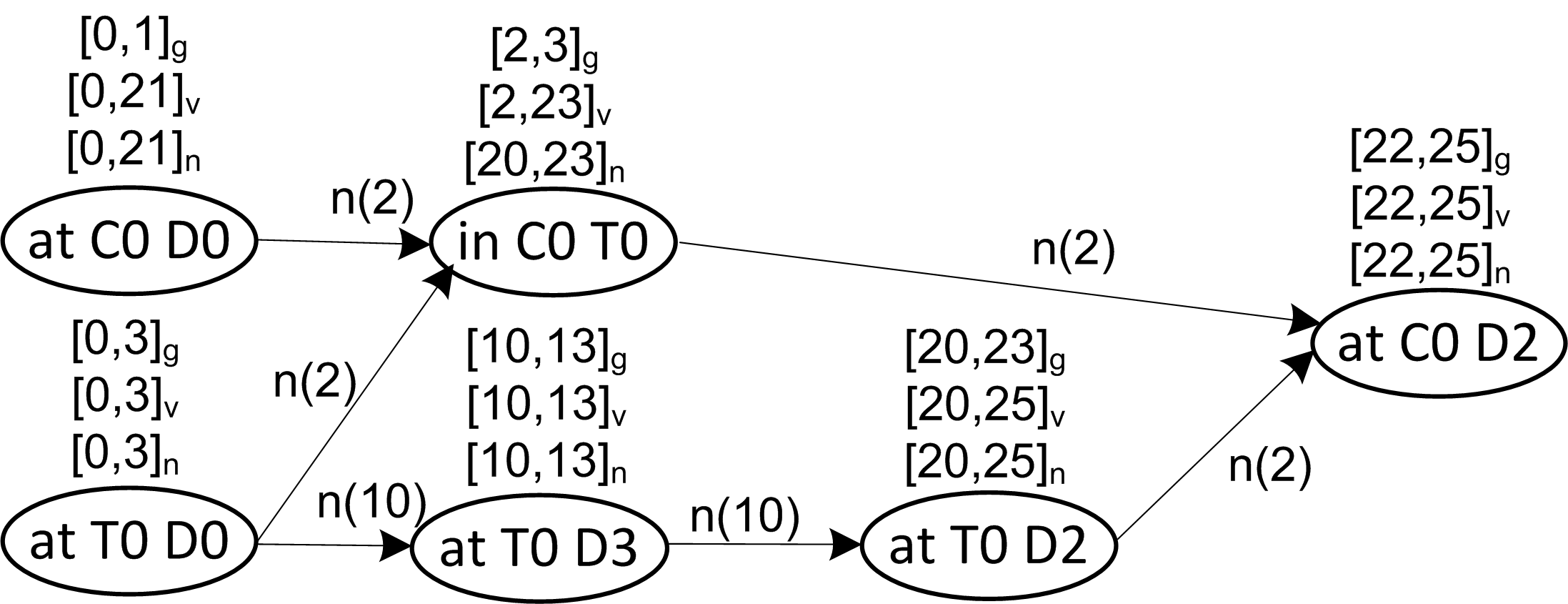, width=8cm}
\caption{TLG for the goal {\small\texttt{(within 25 (at C0 D2))}} after constraint propagation}
\label{fig:TLG_goal25_after}
}
%\end{center}
\end{figure}

In the initial TLG for the goal $g= \texttt{((at C0 D2),25)}$ shown in Figure \ref{fig:TLG_ini_goal25before}, we can observe that nodes are landmarks labeled with the three temporal intervals and edges are labeled with a necessary or dependency ordering constraint plus a temporal distance. For example, $\texttt{(in C0 T0)} \prec_{n(2)} \texttt{(at C0 D2)}$ means it is necessary to have the crate \texttt{C0} into the truck \texttt{T0} at least 2 time units (duration of the action \texttt{unload}) before having the crate \texttt{C0} at \texttt{D2}. In this case, \texttt{(in C0 T0)} is a $SCond$ and \texttt{(at C0 D2)} is an $EEff$ of the same \texttt{unload} action, respectively. Figure \ref{fig:TLG_ini_goal25before} does not picture any dependency ordering because the deadline to have \texttt{C0} at \texttt{D2} is 25, which compels \texttt{T0} to reach \texttt{D2} via \texttt{D3}. However, the TLG for the goal {\small\texttt{(within 40 (at C0 D2))}} shown in Figure \ref{fig:TLG_ini} pictures a dependency ordering $\texttt{(at T0 D0)} \prec_{d(20)} \texttt{(at T0 D2)}$ . The distance 20 denotes that \texttt{T0} must be in \texttt{D0} 20 times units before reaching \texttt{D2}, which is the minimal distance to reach \texttt{D2} from \texttt{D0}. Given that the deadline for the goal is 40 in this case, \texttt{T0} can reach \texttt{D2} through \texttt{D3} or \texttt{D1} but this information is not known yet. This is the reason of the dependency ordering in Figure \ref{fig:TLG_ini}, which means that at least two \texttt{drive} actions are involved in this causal relationship.

We apply an \emph{interval constraint propagation} that restricts the domain of the temporal intervals accordingly to the type of interval and the distance of the causal relationship. The $min$ endpoints of the intervals are propagated forward in time and the $max$ endpoints are propagated backward along time. A causal relationship of the form $l_i\prec_{\{n,d\}} l_j$ between two landmarks $l_i$ and $l_j$ ($l_i$ is required to generate $l_j$) implicitly defines the following interval constraints:

\vspace{0.2cm}
\begin{tabular}{l}
$min_v(l_j) = \max(min_v(l_j), min_v(l_i) + dist(l_i, l_j))$ \\
$max_g(l_i) = \min(max_g(l_i), max_g(l_j) - dist(l_i, l_j))$ \\
\end{tabular}
\vspace{0.15cm}

Thus, the $min_v$ of $l_j$ is subject to the minimum validity of $l_i$ plus the duration of the action(s) that separates both landmarks. Likewise, the latest time when $l_i$ must start in the plan depends on the latest time when $l_j$ is required minus the temporal distance determined by the duration of the action(s) that are needed to generate $l_j$ from $l_i$. Back to the example of Figure \ref{fig:depots} with goal $g= \texttt{((at C0 D2),25)}$, \tool{TempLM} will update the generation intervals of the landmarks as shown in Table \ref{update_intervals}. Note that the order of the $max_g$ propagation goes backwards from \texttt{D2} (the destination depot) through \texttt{D3} to finally reach \texttt{D0}. The final TLG after propagation is shown in Figure \ref{fig:TLG_goal25_after}.

\begin{table}[h]
%\begin{tabular}{|l|l|}
\begin{tabular}{|p{0.28\textwidth}|p{0.16\textwidth}|}
\hline Causal constraint & Interval update  \\
\hline \small $\texttt{(at T0 D2)} \prec_{n(2)} \texttt{(at C0 D2)}$ & \small $max_g\texttt{(at T0 D2)}=25-2=23$\\
\hline \small $\texttt{(at T0 D3)} \prec_{n(10)} \texttt{(at T0 D2)}$ & \small $max_g\texttt{(at T0 D3)}=23-10=13$\\
\hline \small $\texttt{(at T0 D0)} \prec_{n(10)} \texttt{(at T0 D3)}$ & \small $max_g\texttt{(at T0 D0)}=13-10=3$\\
\hline
\end{tabular}
\caption{Update of generation intervals for the goal {\small\texttt{(within 25 (at C0 D2))}}}
\label{update_intervals}
\end{table}

\vspace{0.1cm}

\textbf{\emph{Mutex relationships}}. Given $l_i\prec_{\{d,n\}} l_j$, if $l_i$ and $l_j$ are mutex \cite{blum97} then $l_i$ and $l_j$ cannot overlap in any way. The propagation of the mutex relationships updates $max_v(l_i)$ to ensure $l_i$ does not overlap with $l_j$. Thus, $max_v(l_i)$ is updated to the minimum value among the current validity endpoints of $l_i$ and the latest time when $l_j$ must start in the plan minus the temporal distance between both landmarks landmarks. Particularly:

\vspace{0.1cm}
\begin{tabular}{l}
$max_v(l_i) = \min(max_v(l_i), min_v(l_j), $\\
$\;\;\;\;\;\;\;\;\;\;\;\;\;\;\;\;\;\;\;\;\;\;\;\;\;\;max_g(l_j) - dist(l_i, l_j))$ \\
$max_g(l_i) = \min(max_v(l_i), max_g(l_i))$ \\
\end{tabular}
\vspace{0.1cm}

\begin{table}[h]
%\begin{tabular}{|l|l|}
\begin{tabular}{|p{0.23\textwidth}|p{0.19\textwidth}|}
\hline Mutex landmarks & Interval update  \\
\hline \small $\texttt{((at T0 D3)}, \texttt{(at T0 D2))}$ & \small $max_v\texttt{(at T0 D3)}=\min(25,20,23-10)=13$\\
\hline \small $\texttt{((at T0 D0)},\texttt{(at T0 D3))}$ & \small $max_v\texttt{(at T0 D0)}=\min(25,10,13-10)=3$\\
\hline
\end{tabular}
\caption{Update of validity intervals}
\label{intervals}
\end{table}

\subsection{Search}

\tool{TempLM} searches in the space of \emph{partial plans}. Nodes are represented by a pair $(\Pi,S_t)$, where $\Pi$ is a conflict-free partial plan and $S_t$ is the state reached at time $t=dur(\Pi)$ after executing $\Pi$ in $I$. Nodes are expanded by finding the earliest start time of the set of applicable actions in $S_t$. Each node is associated to a TLG. A newly inserted action may cause the appearance of new temporal landmarks in the TLG of a node and the propagation of the temporal constraints will update the landmarks intervals. As a result of the propagation, if for a given deadline $(p,t)$ it turns out that $max_g(p)>t$ or some inconsistency is found in the endpoints of the landmarks intervals, the node is pruned.

\section{State trajectory constraints in PDDL3.0}
\label{trajectory_constraints}

PDDL3.0 is the language used at the Fifth International Planning Competition (IPC-2006)\footnote{http://www.icaps-conference.org/index.php/Main/Competitions}. This extended language introduces new expressive functionalities such as strong and soft \emph{constraints on plan trajectories} and soft problem goals or \emph{preferences} \cite{GereviniHLSD09}. In this work, we will exclusively focus on the \emph{strong state trajectory constraints} and we will analyze how these are represented, interpreted and handled when using temporal landmarks.

State trajectory constraints are used to express conditions that must be met by the entire sequence of states visited during the execution of a plan. They are expressed through temporal modal operators over first order formulae involving state predicates. Actually, all the constraints expressed with the temporal modal operators of PDDL3.0 specify a temporal interval at which the state predicate must hold along the sequence of states of the plan execution. In the following, we analyze the semantics of the ten modalities of constraints introduced in PDDL3.0 as well as how they would be encoded in a framework based on temporal landmarks.

\subsection{The operator \texttt{at end}}

The syntax of this constraint is \texttt{(at end <GD>)}, where \texttt{<GD>} is a goal descriptor (a first-order logic formula). It is used to identify conditions that must hold in the final state when the plan has been executed, making them equivalent to traditional goal conditions. Whenever a goal condition with no temporal modal operator is specified in a problem file, it is assumed to be an \texttt{(at end)} condition, thus preserving the standard meaning for existing goal specifications. The semantics of this operator is shown in formula \ref{atend}.

\vspace{-0.3cm}
\begin{equation}
\begin{aligned}
& \big<(S_0, 0),(S_1, t_1), \ldots, (S_n, t_n)\big> \models (\texttt{at end} \; \phi) \\
& \textrm{iff} \; S_n \models \phi
\end{aligned}
\label{atend}
\end{equation}

\vspace{0.1cm}

For example, \texttt{(at end (at truck1 cityA))} indicates that \texttt{truck1} must be in \texttt{cityA} at the goal state. The operator \texttt{(at end $\phi$)} defines an interval $[t_i,t_j]$ for the occurrence of the goal condition $\phi$ such that $0 \leq i \leq n$ and $j=n$. Note that the fulfillment of $\phi$ throughout $[t_i,t_j]$ does not necessarily imply that there must be a single occurrence of $\phi$. Particularly, the expression \texttt{(at end $\phi$)} refers to the last appearance of $\phi$ in the plan so that the constraint will be met for such occurrence of $\phi$ as long as $j=n$.

\vspace{0.1cm}

Taking into account the above considerations, a constraint of the form \texttt{(at end $\phi$)} allow us to make the following implications regarding the information of temporal landmarks:

\begin{enumerate}
\item $\phi$ is a temporal landmark since it is a mandatory condition to be satisfied in a particular time interval
\item given that $\phi$ is needed \texttt{at the end}, $max_n(\phi) = t_n$, which implies that $max_v(\phi) = t_n$ because $max_n(\phi) \leq max_v(\phi)$
\item the non-compliance of the constraint in the TLG of a node cannot be used to prune partial plans during search. Only when the plan is complete, the non-compliance of this constraint will be used to discard a plan as a valid solution.
\end{enumerate}

\subsection{The operator \texttt{always}}

The semantics of a constraint \texttt{(always <GD>)} is shown in formula \ref{always}, which expresses that the goal condition  must hold in every state in order for the modal formula to hold over the trajectory.

\vspace{-0.3cm}
\begin{equation}
\begin{aligned}
& \big<(S_0, 0),(S_1, t_1), \ldots, (S_n, t_n)\big> \models (\texttt{always} \; \phi) \\
& \textrm{iff} \; \forall i:0 \leq i \leq n \, \cdot \, S_i \models \phi
\end{aligned}
\label{always}
\end{equation}

\vspace{0.1cm}

A constraint \texttt{(always $\phi$)} expresses that the goal condition $\phi$ must be true throughout the plan. For instance, if the problem requires to have a {\texttt pallet1} clear all the time, we will use the constraint \texttt{(always (clear pallet1))}. Thereby, the expression \texttt{(always $\phi$)} defines a temporal interval $[t_i,t_j]$ for $\phi$ such that $i=0$ and $j=n$. In this case, it is clear that there must be a single occurrence of $\phi$ that holds over $[t_i,t_j]$.

\vspace{0.1cm}

The temporal landmark information that can be inferred through a constraint \texttt{(always $\phi$)} is the following:

\begin{enumerate}
\item $\phi$ is a temporal landmark as the constraint denotes a proposition that must be true in every solution plan over the interval $[0,t_n]$
\item $\phi$ is needed throughout the interval $[0,t_n]$ so $min_n(\phi) = 0$ and $max_n(\phi) = t_n$ , which in turn implies that $min_v(\phi) = 0$ and $max_v(\phi) = t_n$ because $[min_n(\phi), max_n(\phi)] \in [min_v(\phi), max_v(\phi)]$
\item it allows pruning a search node when adding a new action in its TLG entails a modification of the necessity or validity interval of $\phi$.  For example, if the TLG of a node contains \texttt{(always (clear pallet1))} and an action $\alpha = \texttt{(drop P1 T1 pallet1 distributor)}$ is added to the node, then the node will be pruned because the value of $max_v(\texttt{(clear pallet1)})$ is modified since $\alpha$ deletes \texttt{(clear pallet1)}. Additionally, any partial plan of the tree which does not contain $\phi$ will be also pruned.
\end{enumerate}

\subsection{The operator \texttt{at-most-once}}

The syntax of this operator is \texttt{(at-most-once <GD>)} and the semantics is expressed in the formula \ref{at-most-once}.

\vspace{-0.2cm}
\begin{equation}
\begin{aligned}
& \big<(S_0, 0),(S_1, t_1), \ldots, (S_n, t_n)\big> \models (\texttt{at-most-once} \; \phi) \\
& \textrm{iff} \; \forall i:0 \leq i \leq n \, \cdot \\
& \;\;\;\;\;\;\; \textrm{if} \; S_i \models \phi \; \textrm{then} \; \exists j: \; j \geq i \cdot \forall k: i \leq k \leq j \cdot  S_k \models \phi \\
& \;\;\;\;\;\;\; \textrm{and} \;  \forall k: k > j \cdot S_k \models \neg \phi
\end{aligned}
\label{at-most-once}
\end{equation}

A constraint \texttt{(at-most-once $\phi$)} obviously denotes that $\phi$ must occur \texttt{at most once} in the plan, if any. That is, this constraint does not impose a mandatory occurrence of $\phi$ but if it happens then only a single occurrence of $\phi$ must appear in the plan. Consequently, the single occurrence of $\phi$ will be valid over an interval $[t_i,t_j]$, where $0 \leq i \leq n$ and $j \geq i$.

Regarding a temporal landmark representation, a constraint of the form \texttt{(at-most-once $\phi$)} leads to the following derivations:

\begin{enumerate}
\item $\phi$ cannot be labeled as a temporal landmark since a mandatory occurrence is not imposed
\item if $\phi$ is a landmark then we know that $max_g(\phi) \leq t_n$, which indicates that $\phi$ must be obtained before completion of the plan
\item it prevents having more than one occurrence of $\phi$ so any node that violates this condition will be pruned. This has some implications when solving conflicts that involve adding a new occurrence of $\phi$. For example, let's assume that $max_v(\phi)=d$ and $max_n(\phi)=d'$  such that $d' > d$. In this case, a new occurrence of $\phi$ is needed to satisfy the necessity interval. This conflict is solvable in \tool{TempLM} by introducing another instance of $\phi$ ($\phi'$) as long as $\phi'$ is consistent with the intervals of the rest of the landmarks in the TLG of the node \cite{MarzalSO16}. However, such a conflict would be unsolvable if a constraint \texttt{(at-most-once $\phi$)} exists in the planning problem specification.

\end{enumerate}

\subsection{The operator \texttt{sometime}}

The semantics of a constraint \texttt{(sometime <GD>)} is presented in formula \ref{sometime}. As the name and semantics express, a constraint \texttt{(sometime $\phi$)} indicates that $\phi$ must occur \emph{at least once} in the plan. Every single occurrence of $\phi$ must hold over an interval $[t_i,t_j]$, where $0 \leq i \leq n$ and $j \geq i$.

\vspace{-0.2cm}
\begin{equation}
\begin{aligned}
& \big<(S_0, 0),(S_1, t_1), \ldots, (S_n, t_n)\big> \models (\texttt{sometime} \; \phi) \\
& \textrm{iff} \; \exists i:0 \leq i \leq n \, \cdot S_i \models \phi
\end{aligned}
\label{sometime}
\end{equation}

A constraint of the form \texttt{(sometime $\phi$)} allows us to derive the following information related to temporal landmarks:

\begin{enumerate}
\item $\phi$ is a temporal landmark as it must necessarily occur in the plan at least once
\item it must hold $max_g(\phi) \leq t_n$ to ensure that $\phi$ occurs at least in the last state of the plan trajectory
\item similarly to the \texttt{(at-end $\phi$)} constraint, the non-compliance of this constraint in the TLG of a node cannot be used to prune nodes during search. Once the plan construction is finished, we will be able to discard it as a valid solution in case $\phi$ never holds in the plan.
\end{enumerate}

\subsection{The operator \texttt{within}}

The operator \texttt{within} is used to express deadlines. The syntax of this operator is \texttt{(within <num> <GD>)}, where \texttt{<num>} is any numeric literal (in STRIPS domains it will be restricted to integer values) and \texttt{<GD>} has the same meaning as in all the previous operators. The semantics associated to this operator is shown in formula \ref{within}.

\vspace{-0.2cm}
\begin{equation}
\begin{aligned}
& \big<(S_0, 0),(S_1, t_1), \ldots, (S_n, t_n)\big> \models (\texttt{within}\; t \; \phi) \\
& \textrm{iff} \; \exists i:0 \leq i \leq n \cdot \, S_i \models \phi \wedge t_i \leq t
\end{aligned}
\label{within}
\end{equation}

For example, \texttt{(within 10 (at T0 D3))} specifies that truck \texttt{T0} must be in depot \texttt{D3} by time 10 at the latest. The semantics of the operator \texttt{within} does not state the specific occurrence of the goal to which the constraint is applied in case that \texttt{(at T0 D3)} is achieved more than once in the plan. More specifically, the definition states that, if a goal is achieved more than once in the plan, it suffices one appearance of \texttt{(at T0 D3)} to fulfill the \texttt{within} constraint. On the other hand, there is no indication in the semantics that the goal condition must persist until the goal state; that is, the above constraint is satisfied as long as \texttt{(within 10 (at T0 D3))} is met in the plan irrespective of the final location of truck \texttt{T0}. %Thus, we will consider that the constraint \texttt{(within $t$ $\phi$)} refers to the last appearance of $\phi$.

The information of temporal landmarks that can be derived from a constraint \texttt{(within t $\phi$)} (for a particular occurrence of $\phi$) is:

\begin{enumerate}
\item $\phi$ is a temporal landmark as it must necessarily occur in the plan at least once
\item it must always be true that $max_g(\phi) \leq t$
\item given a partial plan $(\Pi,S_{t'})$ such that $t' \geq t$, the node will be pruned if $\phi$ does not hold in $\Pi$
\end{enumerate}

\subsection{Operators \texttt{always-within}, \texttt{sometime-after} and \texttt{sometime-before}}

These three operators share a similar syntax and semantics as they all involve two goal conditions in the constraint. The syntax is as follows: \texttt{(always-within <num> <GD <GD>)}, \texttt{(sometime-after <GD <GD>)} and \texttt{(sometime-before <GD <GD>)}. The constraints only differ in the temporal interval specified for the occurrence of the second goal condition. The semantics of the three operators are shown in formulas \ref{always_within}, \ref{sometime_after} and \ref{sometime_before}.

\vspace{-0.2cm}
\begin{equation}
\begin{aligned}
& \big<(S_0, 0),(S_1, t_1), \ldots, (S_n, t_n)\big> \models (\texttt{always-within}\; t \; \phi \; \psi) \\
& \textrm{iff} \; \forall i:0 \leq i \leq n \; \textrm{if} \; S_i \models \phi \\
& \;\;\;\;\;\;\;\;\;   \textrm{then} \; \exists j: \, i \leq j \leq n \cdot S_j \models \psi \; \textrm{and} \;  t_j - t_i \leq t \\
\end{aligned}
\label{always_within}
\end{equation}

\vspace{-0.3cm}
\begin{equation}
\begin{aligned}
& \big<(S_0, 0),(S_1, t_1), \ldots, (S_n, t_n)\big> \models (\texttt{sometime-after} \; \phi \; \psi) \\
& \textrm{iff} \; \forall i \cdot 0 \leq i \leq n \; \textrm{if} \; S_i \models \phi \\
& \;\;\;\;\;\;\;\;\;   \textrm{then} \; \exists j: \, i \leq j \leq n \cdot S_j \models \psi
\end{aligned}
\label{sometime_after}
\end{equation}

\vspace{-0.3cm}
\begin{equation}
\begin{aligned}
& \big<(S_0, 0),(S_1, t_1), \ldots, (S_n, t_n)\big> \models (\texttt{sometime-before} \; \phi \; \psi) \\
& \textrm{iff} \; \forall i \cdot 0 \leq i \leq n \; \textrm{if} \; S_i \models \phi \\
& \;\;\;\;\;\;\;\;\;   \textrm{then} \; \exists j: \, 0 \leq j < i \cdot S_j \models \psi
\end{aligned}
\label{sometime_before}
\end{equation}

The semantics of the three operators express the following characteristics:

\begin{itemize}
\item the constraints are not restricted to a single occurrence of $\phi$ and $\psi$
\item the constraints apply if and only if $\phi$ occurs in the plan
\item the constraints imply that for every occurrence of $\phi$ there must exist at least one occurrence of $\psi$ that satisfies the corresponding temporal requirement
\item it is not mandatory that every occurrence of $\psi$ meets the constraint as long as there exists at least one occurrence of $\psi$ that does meet the constraint for every $\phi$
\end{itemize}

Specifically, a constraint of the form \texttt{(always-within $t$ $\phi$ $\psi$)} indicates that $\psi$ must hold within $t$ time units from the occurrence of $\phi$. A constraint \texttt{(sometime-before $\phi$ $\psi$)} is met if $\psi$ holds before $\phi$ and a constraint \texttt{(sometime-after $\phi$ $\psi$)} is satisfied if $\psi$ holds after $\phi$.

\vspace{0.1cm}

Regarding the information of temporal landmarks, we can infer the following derivations:

\begin{enumerate}
\item for the three operators: if $\phi$ is a temporal landmark, then $\psi$ is a temporal landmark too as it must necessarily occur in the plan at least once
\item for the operator \texttt{always-within}: it must hold that $\forall\phi \,\, \exists\psi : max_g(\psi) \leq max_g(\phi) + t$. Thus, assuming that $\phi_1$ is the first occurrence of $\phi$, for the remainder occurrences $\phi_i: i > 1$, if $max_g(\phi_i) \leq max_g(\phi_1) + t$ then the same occurrence of $\psi$ will satisfy all $\phi_i$; otherwise, for occurrences $\phi_j: j > 1$ such that $max_g(\phi_j) > max_g(\phi_1) + t$ a different occurrence of $\psi$, say $\psi'$, will be needed to satisfy the constraint of $\phi_j$.
\item for the operator \texttt{sometime-after}, it must hold that $\forall\phi \,\,\exists\psi : max_v(\psi) \geq max_g(\phi)$
\item for the operator \texttt{sometime-before}, it must hold that $\forall\phi \,\, \exists\psi: max_g(\psi) \leq max_g(\phi)$
\item the existence of a constraint \texttt{(always-within $t$ $\phi$ $\psi$)} allows discarding a node $(\Pi,S_{t'})$, $t' > t$, if $\Pi$ contains $\phi$ but not $\psi$
\item a constraint \texttt{(sometime-before $\phi$ $\psi$)} will allow to immediately prune a node which contains $\phi$ but not $\psi$
\item a constraint \texttt{(sometime-after $\phi$ $\psi$)} can only be used to prune nodes that contain finished plans in which $\phi$ holds and $\psi$ does not.
\end{enumerate}

\subsection{The operator \texttt{hold-during}}

The semantics of a constraint \texttt{(hold-during <num> <num> <GD>)} is expressed in formula \ref{hold-during}, indicating that $\phi$ must hold during the interval $[u_1,u_2)$. More particularly, formula \ref{hold-during} explains three cases: when $[u_1,u_2)$ falls entirely within the plan trajectory (first case); when $[u_1,u_2)$ falls partially within the plan trajectory (second case); and when $[u_1,u_2]$ falls outside the plan trajectory (third case).

\vspace{-0.25cm}
\begin{equation}
\begin{aligned}
& \big<(S_0, 0),(S_1, t_1), \ldots, (S_n, t_n)\big> \models (\texttt{hold-during} \; u_1 \; u_2 \; \phi) \\
&  \textrm{iff} \;\;\; \textrm{if} \;\; t_n > u_1 \;\; \textrm{then} \\
& \;\;\;\;\;\;\;\;\forall i \cdot 0 \leq i \leq n \, \cdot \, \textrm{if} \,\; u_1 \leq t_i < u_2 \,\; \textrm{then} \,\; S_i \models \phi, \\
& \;\;\;\;\;\;\;\;\forall j \cdot 0 \leq j < n \, \cdot \, \textrm{if} \,\; t_j \leq u_1 < t_{j+1} \,\; \textrm{then} \,\; S_j \models \phi \\
& \;\;\;\;\;\;\, \textrm{if} \; t_n \leq u_1 \;\, \textrm{then} \;\, S_n \models \phi
\end{aligned}
\label{hold-during}
\end{equation}

A constraint of the form \texttt{(hold-during $u_1$ $u_2$ $\phi$)} allows us to derive the following information related to temporal landmarks:

\begin{enumerate}
\item $\phi$ is a temporal landmark since it must necessarily occur in the plan at least once
\item for the first case, given that it is mandatory for $\phi$ to hold between $u_1$ and $u_2$, we have that $min_n(\phi) \leq u_1$ and $u_2 < max_n(\phi)$; that is, \texttt{(hold-during $u_1$ $u_2$ $\phi$)} determines that $\phi$ is needed at least between $[u_1,u_2)$
\item for the second case, given that it is mandatory for $\phi$ to hold between $u_1$ and $t_n$, we have that $min_n(\phi) \leq u_1$ and $max_n(\phi)=t_n$; that is, \texttt{(hold-during $u_1$ $u_2$ $\phi$)} determines that $\phi$ is needed at least between $[u_1,t_n)$
\item for the third case, given that it is mandatory for $\phi$ to hold at $t_n$, we have that $min_n(\phi)=max_n(\phi)=t_n$; that is, \texttt{(hold-during $u_1$ $u_2$ $\phi$)} determines that $\phi$ is needed at $t_n$
\item the search process will prune any node in which some restriction modifies the necessity interval $[u_1,u_2)$ of $\phi$ (for the first case) or modifies the necessity interval $[u_1,t_n)$ of $\phi$ (for the second case), or it will discard finished plans that do not contain $\phi$ (for the last case)
\end{enumerate}

\subsection{The operator \texttt{hold-after}}

The semantics of \texttt{(hold-after <num> <GD>)} imposes that the goal condition $\phi$ must hold in a state after $t$ time units have elapsed from the initial state at time 0 (see formula \ref{hold_after}). Note that the semantics does not say that $\phi$ must exclusively hold after time $t$ so it could be the case that $\phi$ also holds before $t$. On the other hand, if $t$ is a time later than the finish time of the plan at $t_n$ then $\phi$ must just hold in the last state (second \emph{if} in formula \ref{hold_after}).

\vspace{-0.3cm}
\begin{equation}
\begin{aligned}
& \big<(S_0, 0),(S_1, t_1), \ldots, (S_n, t_n)\big> \models (\texttt{hold-after} \; t \; \phi) \\
&  \textrm{iff} \;\;\; \textrm{if} \;\; t_n > t \;\; \textrm{then}\; \exists i: 0 \leq i \leq n \, \cdot \,  S_i \models \phi \; \textrm{and} \; t_i > t\\
& \;\;\;\;\;\;\, \textrm{if} \; t_n \leq t \;\, \textrm{then} \;\, S_n \models \phi
\end{aligned}
\label{hold_after}
\end{equation}

A constraint of the form \texttt{(hold-after $t$ $\phi$)} allows us to derive the following information related to temporal landmarks:

\begin{enumerate}
\item $\phi$ is a temporal landmark as it must necessarily occur in the plan at least once
\item the constraint $max_v(\phi) \geq t$ must be met%(Sigo sin tenerlo claro, porque esta restricción indica que $\phi$ debe darse al menos una vez después de $t$ pero no dice nada sobre que no deba darse antes de t). [Podría ser \geq t$???]
\item this constraint can only be used to discard finished plans that do not contain $\phi$
\end{enumerate}

\vspace{0.3cm}

Tables \ref{tab:temp-lms} and \ref{tab:temp-deadline} summarize the landmarks derived from the PDDL3.0 modal operators and the updates applied on the endpoints of the landmarks intervals, respectively.

\begin{table}[h]
\begin{tabular}{|l|l|}
\hline Constraint & Landmark  \\
\hline $(\texttt{at end} \; l)$ & $l$ \\
\hline $(\texttt{always} \; l)$ & $l$ \\
\hline $(\texttt{at-most-once} \; l)$ & - \\
\hline $(\texttt{sometime} \; l)$ & $l$ \\
\hline $(\texttt{within} \; t \; l)$ & $l$ \\
\hline $(\texttt{always-within} \; t \; l_i \; l_j)$ & if $l_i$ is a landmark,  \\
& then $l_j$ is a landmark \\
\hline $(\texttt{sometime-before} \; l_i \; l_j)$ & if $l_i$ is a landmark,\\
& then $l_j$ is a landmark  \\
\hline $(\texttt{sometime-after} \; l_i \; l_j)$ & if $l_i$ is a landmark, \\
& then $l_j$ is a landmark  \\
\hline $(\texttt{hold-during} \; t_1 \; t_2 \; l)$ & $l$  \\
\hline $(\texttt{hold-after} \; t \; l)$ & $l$ \\
\hline
\end{tabular}
\caption{Creation of temporal landmarks}
\label{tab:temp-lms}
\end{table}

\begin{table}[h]
\begin{tabular}{|l|l|}
\hline Constraint & Our Model \\
\hline $(\texttt{at end} \; l)$ & $max_n(l)=max_v(l)= t_n$ \\
\hline $(\texttt{always} \; l)$ & $min_n(l)=min_v(l)= t_0$ \\
 & $max_n(l)=max_v(l)= t_n$ \\
\hline $(\texttt{at-most-once} \; l)$  & $max_g(l) \leq t_n$ \\
\hline $(\texttt{sometime} \; l)$ & $max_g(l) \leq t_n$ \\
\hline $(\texttt{within} \; t \; l)$ & $max_g(l) \leq t$ \\
\hline $(\texttt{always-within} \; t \; l_i \; l_j)$ & $max_g(l_j) \leq max_g(l_i) + t$ \\

\hline $(\texttt{sometime-before} \; l_i \; l_j)$ & $max_v(l_j) \geq max_g(l_i)$ \\

\hline $(\texttt{sometime-after} \; l_i \; l_j)$ & $max_g(l_i) \leq max_g(l_j)$ \\

\hline $(\texttt{hold-during} \; t_1 \; t_2 \; l)$  & first case: \\ & $min_n(l) \leq t_1$ $\; max_n(l) > t_2$\\
& second case: \\ & $min_n(l) \leq t_1$ $\; max_n(l) = t_n$\\
& third case: \\ & $min_n(l) = max_n(l) = t_n$\\
\hline $(\texttt{hold-after} \; t \; l)$  & $max_v(l) \geq t$\\
\hline
\end{tabular}
\caption{Temporal constraints on the endpoints of the intervals}
\label{tab:temp-deadline}
\end{table}

\section{Application examples}
\label{example}

In this section we present some practical examples that show the behaviour of \tool{TempLM} when handling several PDDL3.0 state trajectory constraints on the scenario introduced in Figure \ref{fig:depots}. The actions of this domain are:

\begin{scriptsize}
\begin{verbatim}
(:durative-action drive
  :parameters (?truck - truck ?loc-from - place
               ?loc-to - place ?driver - driver)
  :duration (= ?duration (time-to-drive ?loc-from ?loc-to))
  :condition (and (at start (at ?truck ?loc-from))
                  (at start (link ?loc-from ?loc-to)))
  :effect (and (at start (not (at ?truck ?loc-from)))
               (at end (at ?truck ?loc-to))))
\end{verbatim}
\end{scriptsize}

The \texttt{drive} action allows a \texttt{?truck} to move between two locations \texttt{?loc-from} and \texttt{?loc-to}, which are a depot or a distributor. The truck can move without carrying any crates. The duration of this action is given by the time to drive between the two locations.

\begin{scriptsize}
\begin{verbatim}
(:durative-action load
  :parameters (?obj - crate ?truck - truck
               ?surf - surface ?loc - place)
  :duration (= ?duration 2)
  :condition (and
   (at start (at ?obj ?loc))(at start (on ?obj ?surf))
   (over all (at ?truck ?loc))(at start (clear ?obj))
   (over all (at ?surf ?loc)))
  :effect (and
   (at start (not (at ?obj ?loc)))(at end (clear ?surf))
   (at start (not (on ?y ?z)))(at end (in ?obj ?truck))))
\end{verbatim}
\end{scriptsize}

The \texttt{load} action is used to load a crate \texttt{?obj}, which is onto the surface \texttt{?surf} of the location \texttt{?loc}, into the \texttt{?truck}. As a side effect, the surface where the crate was found is cleared.

\begin{scriptsize}
\begin{verbatim}
(:durative-action unload
  :parameters (?obj - crate ?truck - truck
               ?surf - surface ?loc - place)
  :duration (= ?duration 2)
  :condition (and
   (over all (at ?truck ?loc))(at start (in ?obj ?truck))
   (over all (at ?surface ?loc))(at start (clear ?surf)))
  :effect (and
   (at start (not (in ?obj ?truck)))(at end (at ?obj ?loc))
   (at start (not (clear ?surf))) (at end (on ?obj ?surf))))
\end{verbatim}
\end{scriptsize}

The \texttt{unload} action puts a crate \texttt{?obj} which is into a \texttt{?truck} onto a surface \texttt{?surf} located at the same place \texttt{?loc} than the \texttt{?truck}.

We will now show the temporal information that can be extracted when applying several state trajectory constraints on this problem.

\vspace{0.15cm}

\textbf{Example with a \texttt{within} constraint}. Let's assume the problem goal is \texttt{(within 20 (at C0 D2))}. In this case, \tool{TempLM} finds that $min_g(\texttt{at C0 D2})=22$ (see the calculation of this value in page 2) and $max_g(\texttt{at C0 D2})=20$ so the TLG will not be generated and \tool{TempLM} will return 'unsolvable problem'.

\begin{figure*}
%\begin{center}
{\centering
\epsfig{file=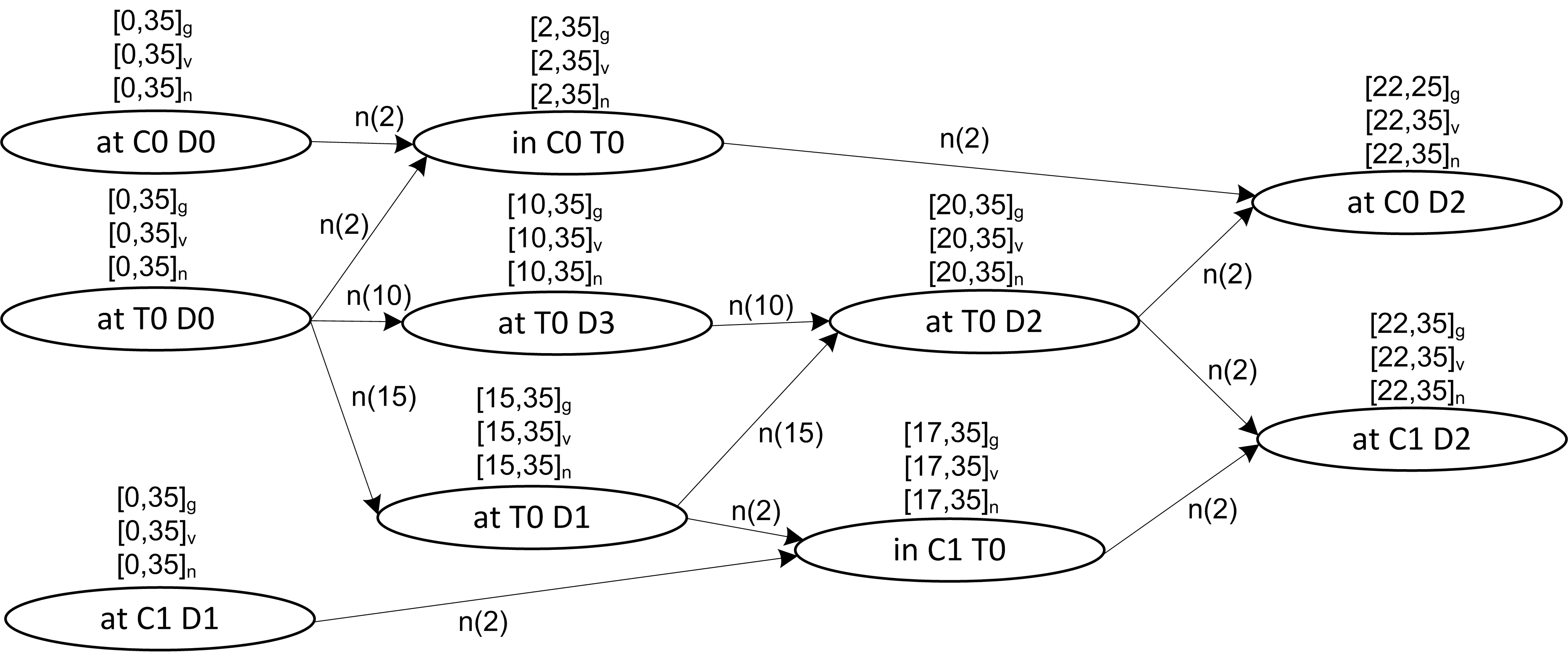, width=14cm}
\caption{Initial TLG for the goals \texttt{(within 25 (at C0 D2))} and \texttt{(within 35 (at C1 D2))} before propagation}
\label{fig:TLGWithin}
}
%\end{center}
\end{figure*}

Let's now suppose that the problem goals are \texttt{(within 25 (at C0 D2))} and \texttt{(within 35 (at C1 D2))}. The initial TLG before propagation for this problem is shown in Figure \ref{fig:TLGWithin}\footnote{Only the most relevant landmarks are displayed}. As explained in section \emph{Overview of \tool{TempLM}}, \texttt{(at T0 D3)} is a landmark because it is the only way to satisfy \texttt{(within 25 (at C0 D2))}. On the other hand, \texttt{(at T0 D1)} is a landmark too because \texttt{T0} must go by distributor \texttt{D1} to load crate \texttt{C1}. Then, the min endpoint of the validity interval of \texttt{(at T0 D2)} is updated to: $min_v(\texttt{at T0 D2})=\max(min_v(\texttt{at T0 D2}),min_v(\texttt{at T0 D3})+10, min_v(\texttt{at T0 D1)})+15)=30$. After propagating this interval modification, we have that $min_v(\texttt{at C0 D2})=32$ , which obviously entails an interval inconsistency because $max_g(\texttt{at C0 D2})=25 < min_v(\texttt{at C0 D2})=32$. This is an indication that the only way of achieving \texttt{(at C0 D2)} at time 25 is traveling through distributor \texttt{D3}. In turn, this means that \texttt{(at T0 D1)} is ordered after \texttt{(at T0 D2)} (i.e., \texttt{C1} is transported after \texttt{C0}), thus causing another inconsistency in the landmark \texttt{(at C1 D2)} because \texttt{C1} will not be delivered in time. This is also an indication that the problem is unsolvable and \tool{TempLM} will detect this situation before even starting the search process.

\begin{figure}
%\begin{center}
{\centering
\epsfig{file=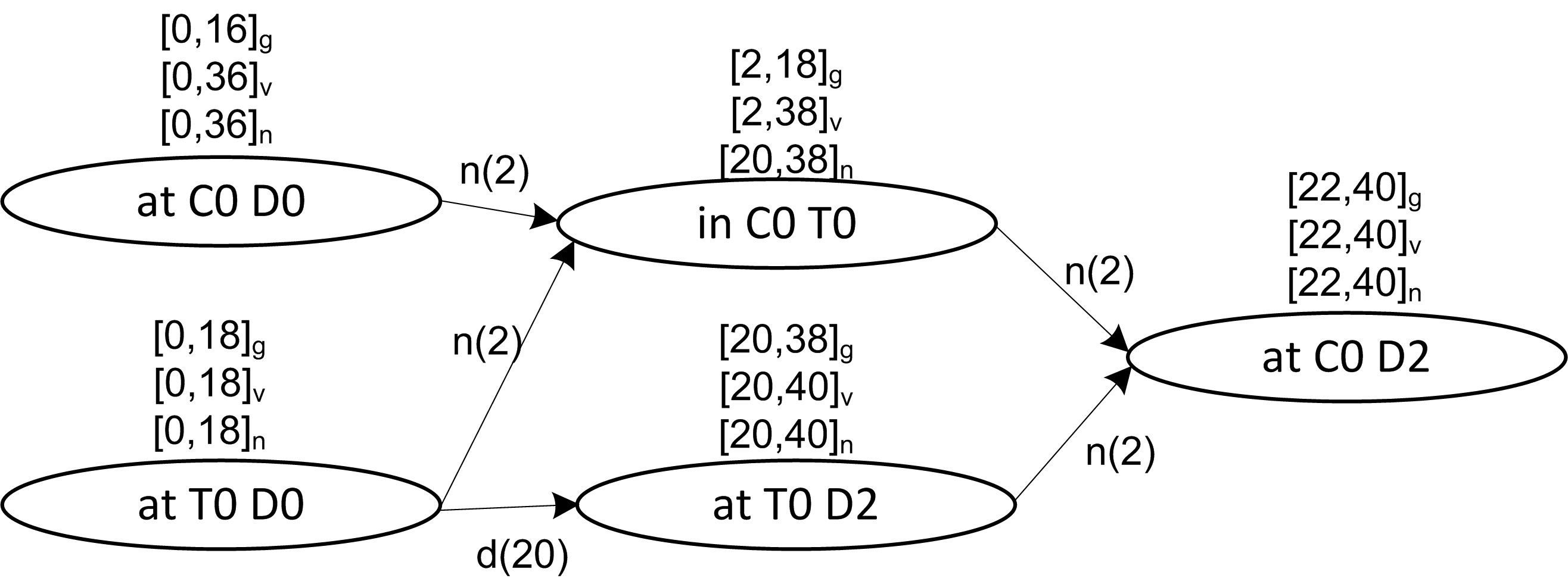, width=8cm}
\caption{Initial TLG for the goal {\small\texttt{(within 40 (at C0 D2))}}}
\label{fig:TLG_ini}
}
%\end{center}
\end{figure}

\vspace{0.15cm}
\textbf{Example with an \texttt{always} constraint}. In this case, we will assume that the only goal is \texttt{(within 40 (at C0 D2))}. Figure \ref{fig:TLG_ini} shows the initial TLG for this goal\footnote{The landmarks of the initial state are not shown for the sake of simplicity}. In the figure, we can observe a dependency ordering between \texttt{(at T0 D0)} and \texttt{(at T0 D2)}, representing that \texttt{T0} will reach \texttt{D2} after \texttt{D0} (in this case, since the deadline is at 40 we don't know yet whether the route of \texttt{T0} to reach \texttt{D2} must go through \texttt{D1} or \texttt{D3}). Another observation is that the pallet on which \texttt{C0} will be stacked is unknown, reason why \texttt{(on C0 P2)} and \texttt{(on C0 P3)} are not \emph{landmarks}.

If we define \texttt{(always (clear P2))}, \texttt{(clear P2)} becomes a landmark with validity interval $[0,40]$. During the search process, two actions that achieve the effect \texttt{(at C0 D2)} are found: \texttt{(unload C0 T0 P2 D2)} and \texttt{(unload C0 T0 P3 D2)}. Given that the application of \texttt{(unload C0 T0 P2 D2)} modifies the value of $max_v(\texttt{at C0 D2})$ when crate \texttt{C0} is unloaded in \texttt{P2}, and that an \texttt{always} constraints compels  $max_v(\texttt{at C0 D2})=t_n$ throughout the plan, the only viable option is to use the action \texttt{(unload C0 T0 P3 D2)} and \tool{TempLM} would discard the node that unloads \texttt{C0} in \texttt{P2}.

\vspace{0.15cm}

\textbf{Example with an \texttt{always-within} constraint}. Following with a problem that contains the single goal \texttt{(within 40 (at C0 D2))}, let's suppose that we add \texttt{(always-within 22 (in C0 T0) (at C0 D2))}. Since the deadline for the problem goal \texttt{(at C0 D2)} is not very tight ($max_g(\texttt{at C0 D2})=40$), the new constraint does not affect the max endpoint of the generation interval of the goal. However, new information could be inferred during the search process. For instance, if \texttt{(in C0 T0)} is achieved at $t=5$ then $max_g(\texttt{at C0 D2})=27$, which would allow us to infer that \texttt{(at T0 D3)} must be now a landmark.

\vspace{0.15cm}

\textbf{Example with a \texttt{hold-during} constraint}. Assuming we have the same goal as above \texttt{(within 40 (at C0 D2))}, let's suppose the truck \texttt{T0} must go through some maintenance repair in depot \texttt{D0} before starting the delivery. We define the restriction \texttt{(hold-during 0 10 (at T0 D0))} to denote that \texttt{T0} must stay at \texttt{D0} for 10 time units for the maintenance work. This restriction does not alter the initial necessity interval of \texttt{(at T0 D0)}, which is $[0,18]$ as can be seen in Figure \ref{fig:TLG_ini} (18 is the latest time that \texttt{T0} can stay in \texttt{D0} in order to achieve the goal at 40). Nodes that comprise partial plans in which \texttt{T0} is not in \texttt{D0} up to time 10 will be eliminated during the search process; that is, nodes that include a drive action \texttt{(drive T0 D0 X)} between 0 and 10.

\vspace{0.15cm}
\textbf{Example with an \texttt{at end} constraint}.  Assuming we have the same goal as above, \texttt{(within 40 (at C0 D2))}, this examples shows a situation in which besides satisfying the goal, the truck \texttt{T0} must end the transportation at distributor \texttt{D3}. This implies defining also the constraint \texttt{(at end (at T0 D3))}, which makes \texttt{(at T0 D3)} become a landmark with validity interval $[0,t_n]$, $max_n \texttt{(at T0 D3)}=t_n$ and introduces the ordering $\texttt{(at T0 D2)} \prec_d \texttt{(at T0 D3)}$. This ordering is motivated because \texttt{(at T0 D2)} and \texttt{(at T0 D3)} are mutex and \texttt{(at T0 D3)} must happen at the end due to the constraint.

During the plan construction, given that the goal deadline is at time 40 and hence truck \texttt{T0} can reach distributor \texttt{D2} either traversing \texttt{D1} or \texttt{D3}, the search tree will comprise two branches that follow these two alternatives. Let's analyze the impact of constraint \texttt{(at end (at T0 D3))} in the second branch, the one that traverses \texttt{D3}. In this case, \texttt{T0} must go through \texttt{D3} to reach \texttt{D2}, which implies $\texttt{(at T0 D3)} \prec_n \texttt{(at T0 D2)}$ and $max_v \texttt{(at T0 D3)} < t_n$. Then, a conflict arises because the TLG contains the landmark \texttt{(at T0 D3)} with $max_n \texttt{(at T0 D3)}=t_n$. As explained in the section of the \texttt{at-most-once} modal operator, when a landmark is found to be needed beyond its maximum validity, \tool{TempLM} solves this conflict by introducing a new occurrence of the landmark \texttt{(at T0 D3)}, and this new occurrence is the one that will be ordered before \texttt{(at T0 D2)}. Thus, the final TLG will contain $\texttt{(at T0 D3)} \prec_n \texttt{(at T0 D2)} \prec_d \texttt{(at T0 D3)}$.

\begin{figure*}
%\begin{center}
{\centering
\epsfig{file=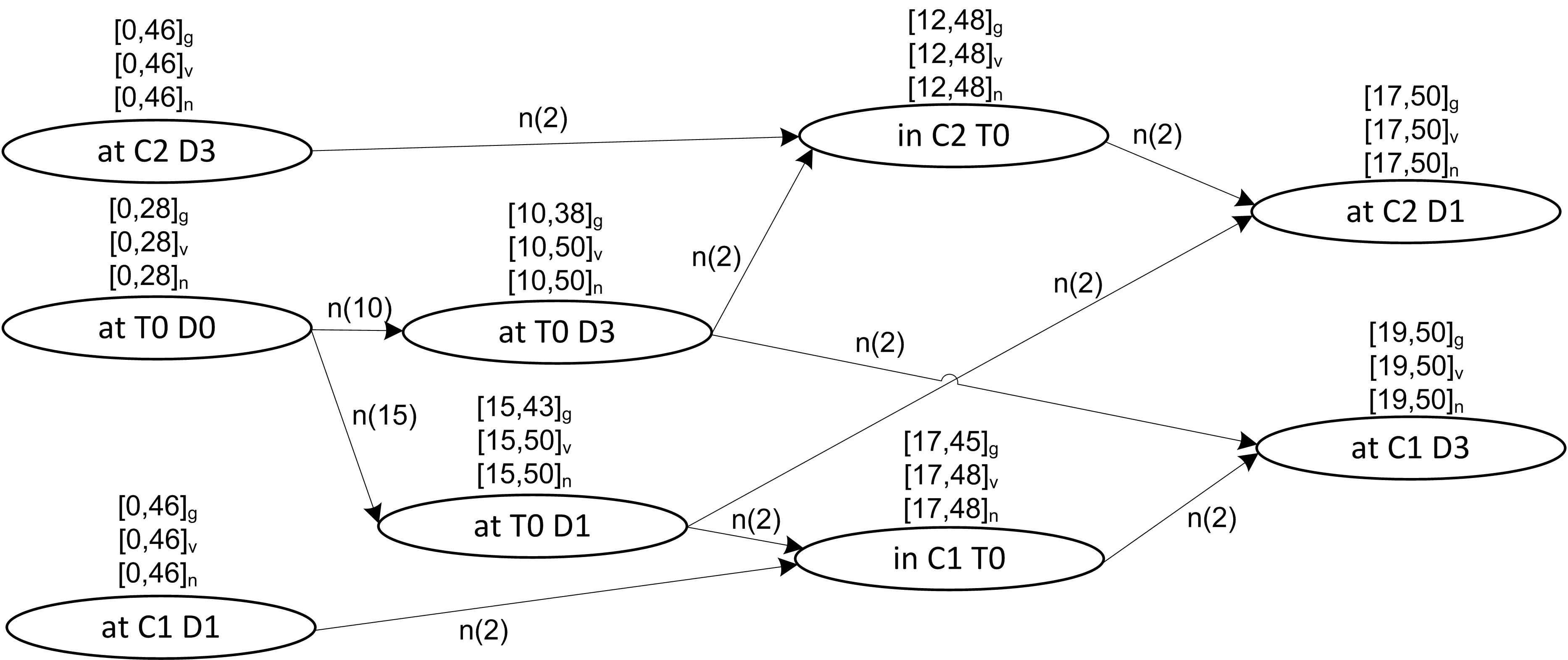, width=14cm}
\caption{Initial TLG for the goal \texttt{(within 50 (at C1 D3))} and \texttt{(within 50 (at C2 D1))}}
\label{fig:TLGAtMost}
}
%\end{center}
\end{figure*}

\vspace{0.15cm}
\textbf{Example with an \texttt{at-most-once} constraint}. In this example, the goal is to switch the location of two packages between distributors: \texttt{(within 50 (at C1 D3))} and \texttt{(within 50 (at C2 D1))}. Figure \ref{fig:TLGAtMost} shows the initial TLG for this problem. We can observe there is an inconsistency between the validity intervals of \texttt{(at T0 D3)} and \texttt{(at T0 D1)} because these two landmarks are mutex and cannot co-exist (the truck \texttt{T0} cannot be simultaneously in distributor \texttt{D1} and distributor \texttt{D3}). \tool{TempLM} is not able to decide the order of these two landmarks with the current deadlines, being thus possible $\texttt{(at T0 D3)} \prec_d \texttt{(at T0 D1)}$ or $\texttt{(at T0 D1)} \prec_d \texttt{(at T0 D3)}$.

Let's suppose the problem includes now the constraint \texttt{(at-most-once (at T0 D3))} and that a node $n=(\Pi,S_t)$ which contains the ordering $\texttt{(at T0 D3)} \prec_d \texttt{(at T0 D1)}$  in $\Pi$ is found during the search process. $\Pi$ embodies a plan where \texttt{T0} drops first by \texttt{D3} to load crate \texttt{C2}, which in turn implies that \texttt{T0} will need to get back to \texttt{D3} to unload \texttt{C1}, thus violating the constraint \texttt{(at-most-once (at T0 D3))}. Therefore, the node $n$ will be discarded. In this example, the only feasible solution is a plan that contains $\texttt{(at T0 D1)} \prec_d \texttt{(at T0 D3)}$, meaning that by the time \texttt{T0} reaches distributor \texttt{D3} to load crate \texttt{C2}, the truck already contains the crate \texttt{C1} to be unloaded in \texttt{D3}.

\section{Discussion: beyond PDDL3.0}

The exposition presented in the two previous sections reveal that the temporal landmarks formalism of \tool{TempLM} is a very appropriate mechanism to deal with state trajectory constraints. It is certainly true that the functioning of \tool{TempLM} is conditioned to the upper time bound of the plan $T_{\Pi}$, which can be set as the maximum value of all the deadlines constraints defined in the problem or as any particular value, and that the less restrictive $T_{\Pi}$ is, the less information will be extracted from the trajectory constraints. Nevertheless, considering that constraints \texttt{at end}, \texttt{sometime}, \texttt{sometime-after} and \texttt{hold-after} are only applicable over finished plans, and that constraints \texttt{sometime-before}, \texttt{at-most-once} and \texttt{always} are easily checkable in any partial plan regardless the deadlines of the problem, we can conclude that the constraints that mostly affect the behaviour and performance of a temporal planner are \texttt{within}, \texttt{always-within} and \texttt{hold-during}, which all define a deadline constraint. Interestingly, adapting makespan-minimization heuristics to account for state trajectory constraints is still a challenging and unexplored line of investigation.

Besides the potential of temporal landmarks to handle trajectory constraints, we envision some further functionalities. For instance, one is not allowed to express in PDDL3.0 that a proposition $\psi$ must hold within $t$ time units from the end of another proposition $\phi$. The specification of state trajectory constraints that involve two propositions $\phi$ and $\psi$ is always related to the occurrence time of the first proposition $\phi$, irrespective of $\phi$ is true or not when $\psi$ holds. Handling a constraint of the form "$\psi$ must \texttt{hold within} $t$ time units from the end of $\phi$" will be easily encoded with the temporal constraint $max_g(\psi) \leq max_v(\phi) + t$.

Another interesting issue is to be able to specify \emph{persistence} of facts. Persistence would be expressed with \texttt{(within <num> (always <GD>))}, which requires nesting of the modalities and this is not allowed in standard PDDL3.0
syntax\footnote{Personal communication with Derek Long}. If PDDL3.0 were extended to include, for example, a modal operator like \texttt{(persistence $t$ $\phi$)}, this would be easily encoded in \tool{TempLM} as $max_n(\phi) \geq max_g(\phi) + t$.

Last but not least, \tool{TempLM} can also be adapted to the particular features of any temporal model; e.g., Allen's interval algebra \cite{allen83}. Intervals of the algebra would be represented by means of the landmarks intervals and the 13 base relations would be captured by setting the appropriate temporal constraints between the $max_g$ and $max_v$ of the temporal landmarks. For instance, \texttt{(overlaps $\phi$ $\psi$)} would be encoded as $max_v(\phi) \geq max_g(\psi)$; and \texttt{(during $\phi$ $\psi$)} as $max_g(\phi) \geq max_g(\psi)$ and $max_v(\phi) \leq max_v(\psi)$.

A practical application of state trajectory constraints is the delivery of perishable goods such as fish or seafood.  Companies must not only meet the delivery deadlines but also consider the best transport means for each product. Hence, depending on the type of product (fresh, frozen or long-term preserving fish products) and the temperature of the refrigerated transport (ice-cooled or machine-cooled wagons), the amount of time goods are exposed to particular temperatures must not exceed a time limit so as to ensure freshness, nutritional value and food preservation of the fishing goods.

All in all, we can conclude that the temporal formalism of \tool{TempLM} offers a great flexibility to express any kind of temporal constraints in temporal planning problems.

\section{Acknowledgements}

This work has been partly supported by the Spanish MINECO under project TIN2014-55637-C2-2-R and the Valenciam project PROMETEO II/2013/019.

%\newpage

%% References


\bibliographystyle{aaai}
\bibliography{../biblio/tesis_long,../biblio/laura_bib-eva_long}

\end{document}